\documentclass{article}

\usepackage{arxiv}
\usepackage[utf8]{inputenc} % allow utf-8 input
\usepackage[T1]{fontenc}    % use 8-bit T1 fonts
\usepackage{hyperref}       % hyperlinks
\usepackage{url}            % simple URL typesetting
\usepackage{booktabs}       % professional-quality tables
\usepackage{amsfonts}       % blackboard math symbols
\usepackage{nicefrac}       % compact symbols for 1/2, etc.
\usepackage{microtype}      % microtypography
\usepackage{lipsum}
\usepackage{graphicx}
\usepackage{subfigure}
\usepackage{color}
\usepackage{listings}
\usepackage{xcolor}
\usepackage{xparse}
\usepackage{tablefootnote}
\usepackage[inline]{enumitem}  % for inline enumerate

\NewDocumentCommand{\codeword}{v}{%
\texttt{\textcolor{blue}{#1}}%
}

\definecolor{codegreen}{rgb}{0,0.6,0}
\definecolor{codegray}{rgb}{0.5,0.5,0.5}
\definecolor{codepurple}{rgb}{0.58,0,0.82}
\definecolor{backcolour}{rgb}{0.95,0.95,0.92}

\lstdefinestyle{mystyle}{
  backgroundcolor=\color{backcolour}, commentstyle=\color{codegreen},
  keywordstyle=\color{magenta},
  numberstyle=\footnotesize\color{codegray}\ttfamily,
  stringstyle=\color{codepurple},
  basicstyle=\ttfamily\footnotesize,
  breakatwhitespace=false,         
  breaklines=true,                 
  captionpos=b,                    
  keepspaces=true,                 
  numbers=left,                    
  numbersep=5pt,                  
  showspaces=false,                
  showstringspaces=false,
  showtabs=false,                  
  tabsize=2
}

\graphicspath{ {./images/} }

\title{FedLab: A Flexible Federated Learning Framework}

\author{
Dun Zeng$^{1*}$, Siqi Liang$^{2}$\thanks{ Equal contribution.}, Xiangjing Hu$^{3}$, Hui Wang$^{4}$, Zenglin Xu$^{3,4}$\thanks{ Corresponding author.}\\
${^1}$School of Computer Science and Engineering, University of Electronic Science and Technology of China\\
${^2}$Rich Media Big Data Analytics and Application Key Laboratory, \\Shenzhen Research Institute, The Chinese University of Hong Kong\\ 
${^3}$School of Computer Science and Technology, Harbin Institute of Technology Shenzhen\\
${^4}$Peng Cheng Laboratory, Shenzhen, China\\
zengdun@std.uestc.edu.cn, zszxlsq@gmail.com, xiangjinghu@stu.hit.edu.cn, wangh06@pcl.ac.cn, xuzenglin@hit.edu.cn
}

\begin{document}
\maketitle
\begin{abstract}
Federated learning (FL) is a machine learning field in which researchers try to facilitate model learning process among multiparty without violating privacy protection regulations. Considerable effort has been invested in FL optimization and communication related researches. In this work, we introduce \texttt{FedLab}, a lightweight open-source framework for FL simulation. The design of \texttt{FedLab} focuses on FL algorithm effectiveness and communication efficiency. Also, \texttt{FedLab} is scalable in different deployment scenario. We hope \texttt{FedLab} could provide flexible API as well as reliable baseline implementations, and relieve the burden of implementing novel approaches for researchers in FL community. The source code is available at \codeword{https://github.com/SMILELab-FL/FedLab}.

\end{abstract}

% keywords can be removed

\section{Introduction}
Federated learning (FL), proposed by Google at the very beginning \cite{DBLP:conf/aistats/McMahanMRHA17}, is recently a burgeoning research area of machine learning, which aims to protect individual data privacy in distributed machine learning process, especially in finance \cite{DBLP:journals/corr/abs-2010-05867}, smart healthcare \cite{DBLP:journals/jhir/XuGSWBW21, DBLP:journals/ijmi/BrisimiCMOPS18} and edge computing \cite{DBLP:journals/corr/abs-1909-12326, DBLP:journals/corr/abs-2106-05223}. Different from traditional data-centered distributed machine learning, participants in FL setting utilize localized data to train local model, then leverages specific strategies with other participants to acquire the final model collaboratively, avoiding direct data sharing behavior.\par

\begin{figure}
    \centering
    \includegraphics[width=0.8\textwidth]{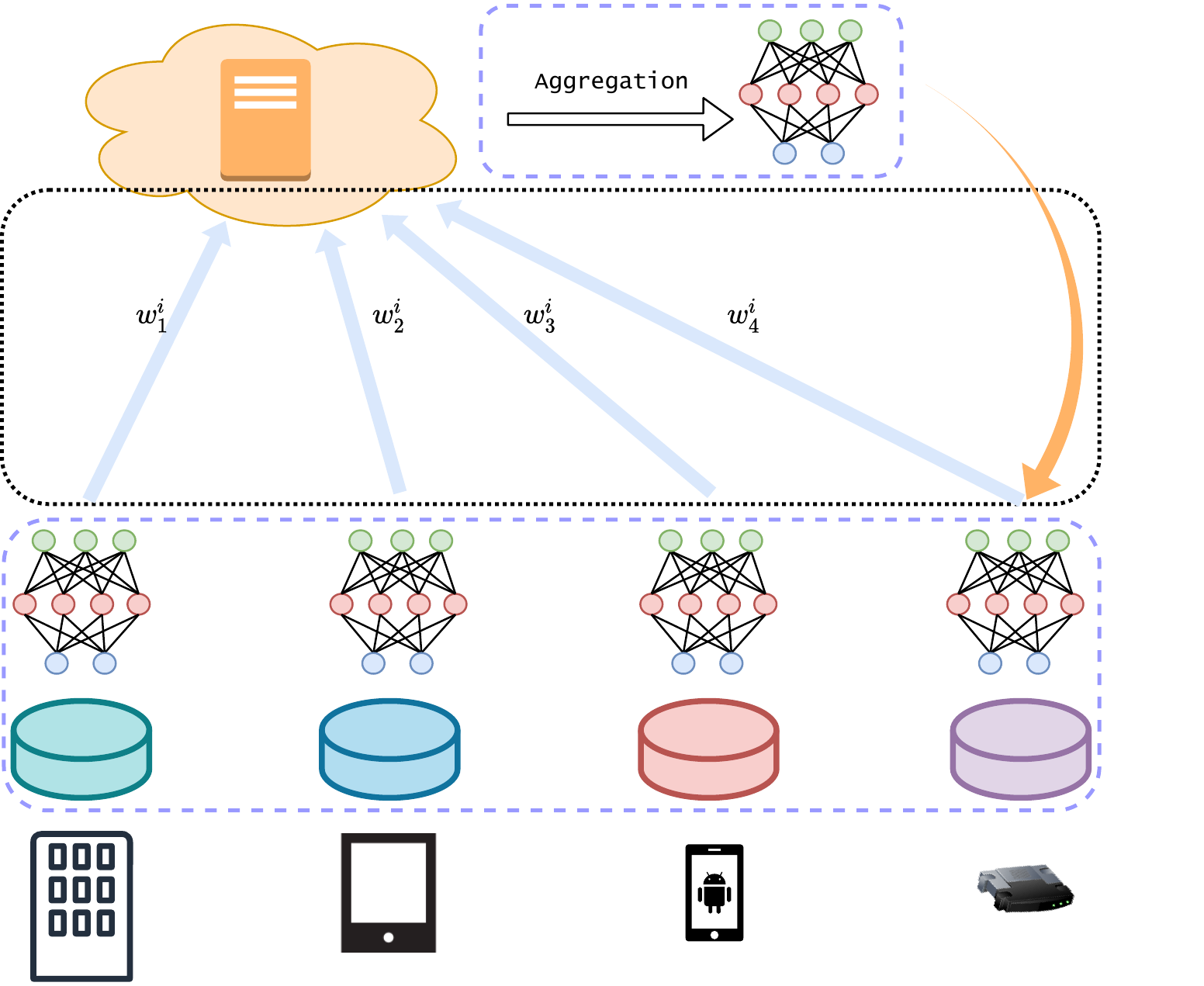}
    \caption{The paradigm of Federated Learning. The content in black dashed box indicates the communication strategy of FL system. The content in blue dashed box indicates the FL optimization, including global aggregation and local optimization.}
    \label{fig:fl_paradigm}
\end{figure}

Though it might differ in specific methodologies, current FL schemes can be summarized as repetition of training rounds, with each integrated by several basic steps: 
\begin{enumerate*}[label=\itshape\roman*\upshape)]
    \item \label{step1} local update on client's model using their own localized data;
    \item \label{step2} clients upload their local trained model parameters to server;
    \item \label{step3} server performs aggregation strategy on collected clients' model parameters to obtain global model;
    \item \label{step4} server selects a subset of clients and distributes the latest global model to them.
\end{enumerate*}
Many FL researches try to improve algorithm effectiveness or efficiency on only one or more steps in this workflow with different scenarios: \cite{DBLP:conf/mlsys/LiSZSTS20} suggests to add regularization term in step \ref{step1} to achieve more robust convergence in heterogeneous settings; \cite{lin2017deep} applies gradient compression method in step \ref{step2} to reduce communication bandwidth; \cite{shokri2015privacy} tries to modify in step \ref{step1}, \ref{step2} and \ref{step3} for privacy-preserving purpose; \cite{sattler2020clustered} proposes better sample strategy in step \ref{step4} to address suboptimal result problem in Federated Multi-Task Learning. These indicate that the implementation of many FL algorithms only requires modification on several components of common workflow, without the necessity of repetitive implementation on basic FL workflow. The paradigm of FL and related research points are as depicted in figure \ref{fig:fl_paradigm}.

However, though with several FL related frameworks or platforms available, researchers still prefer to implement FL algorithms using PyTorch \cite{paszke2019pytorch} or TensorFlow \cite{abadi2016tensorflow} from scratch \cite{jeong2021fedmatchcode, tianli2021fedproxcode}. This inefficient modus operandi in FL community can hamper researchers' enthusiasm in both procedures of reproducing previous work and fast verification of new ideas.  \par

\begin{table}[]
\centering
\begin{tabular}{cccccc}
\hline
Method & Step \ref{step1} & Step \ref{step2} & Step \ref{step3} & Step \ref{step4} & Platform \\ \hline
% \cite{DBLP:conf/iclr/LiSBS20}  &       &   \checkmark    &   \checkmark    &   \checkmark    &       TensorFlow        \\
\cite{DBLP:conf/iclr/WangYSPK20}       &       &   \checkmark    &   \checkmark    &   \checkmark    &   PyTorch\tablefootnote{Official code: \url{https://github.com/IBM/FedMA}.}      \\
\cite{DBLP:conf/nips/GhoshCYR20}       &   \checkmark    &       &       &  \checkmark     &       TensorFlow\tablefootnote{Official code: \url{https://github.com/jichan3751/ifca}.}        \\
\cite{DBLP:conf/nips/DinhTN20}       &      &   \checkmark    &       &   \checkmark    &       PyTorch\tablefootnote{Official code: \url{https://github.com/CharlieDinh/pFedMe}.}      \\
\cite{DBLP:conf/iclr/LiJZKD21}       &       &  \checkmark     &       &   \checkmark    &      PyTorch\tablefootnote{Official code: \url{https://github.com/med-air/FedBN}.}         \\ 
\cite{DBLP:conf/iclr/AcarZNMWS21}       &       &     \checkmark      &    \checkmark      &   \checkmark    &      PyTorch\tablefootnote{Official code: \url{https://github.com/alpemreacar/FedDyn}.}         \\ 
\cite{NEURIPS2021_64be20f6}       &       &           &          &   \checkmark    &       PyTorch\tablefootnote{Official code: \url{https://github.com/hmgxr128/MIFA_code}}        \\  
\cite{NEURIPS2021_4c27cea8}       &   \checkmark    &    \checkmark    &          &   \checkmark    &       Sklearn\tablefootnote{Official code: \url{https://github.com/daizhongxiang/Differentially-Private-Federated-Bayesian-Optimization}.}        \\  
\cite{oh2022fedbabu}       &       &   \checkmark    &   \checkmark    &    \checkmark   &      PyTorch\tablefootnote{Official code: \url{https://github.com/jhoon-oh/fedbabu}.}         \\ \hline
\end{tabular}
\caption{The investigation results of recently published FL algorithms.}
\end{table}

To relieve the burden of researchers in implementing FL algorithms and emancipate FL scientists from repetitive implementation of basic FL setting, we introduce highly customizable framework \texttt{FedLab} in this paper. \texttt{FedLab} provides the necessary modules for FL simulation, including communication, compression, model optimization, data partition and other functional modules. \texttt{FedLab} users can build FL simulation environment with custom modules like playing with LEGO bricks. In all, we make the following contributions to FL community:
\begin{itemize}
    \item A flexible FL framework \texttt{FedLab} is proposed, in which the flexibility is given by highly customizable interfaces and scalability in FL system. \texttt{FedLab} allows users focus on interested components design while keeping other part default. What's more, \texttt{FedLab} also supports \emph{standalone}, \emph{cross machine} and \emph{hierarchical} simulation paradigms.
    \item Various data partition tools for comprehensive data distribution scenarios in FL. \texttt{FedLab} provides a series of data partition functions as well as built-in data partition schemes for different data distributions over federation. 
    \item Standardized FL implementation schemes are presented through \texttt{FedLab}. For instance, standard synchronous and asynchronous FL system are available. Besides, we also provides FL datasets benchmarks and functional modules for standard FL simulation. 
    \item An open-source group is founded in GitHub repository for \texttt{FedLab}'s continuous
    maintenance. Elaborate document is published as well.
\end{itemize} 

% \textcolor[rgb]{1,0,0}{
% \begin{itemize}
%     \item \textbf{Flexible FL framework.} \texttt{FedLab} is a flexible, modular framework for FL system simulation, in which the flexibility is given by highly customizable interfaces and scalability in FL system. \texttt{FedLab} allows users focus on interested components design while keep other part default. What's more, \emph{Standalone}, \emph{cross machine} and \emph{hierarchical} scenarios are supported by \texttt{FedLab}.
%     \item \textbf{Standard FL implementation.} \texttt{FedLab} is carefully designed to provide standard FL implementation scheme. For instance, standard synchronous and asynchronous communication patterns are available. 
%     \item \textbf{Open-source.}  
%     open-source, welcomes new contributions from other researchers
%     \item \textbf{Comprehensive documents.} User friendly blah blah
% \end{itemize}}

\section{Background}
Current FL community focuses mainly on two major challenges. Firstly, data heterogeneity across clients slows down model convergence \cite{li2019convergence} compared with that of data-center distributed learning \cite{dean2012large}. The other major challenge is communication cost during both model uploading and downloading processes, which is also the bottleneck of distributed learning. There is an urgent need for improvement on communication, especially when it comes to cross-device scenario. A lot of works have been proposed to tackle these two challenges, which can be categorized into optimization algorithms and communication efficient strategies. In this section, these two popular research sub-fields of FL will be further illustrated, and the need of a convenient FL framework suitable for optimization effectiveness and communication efficiency research will be revealed.

\subsection{Optimization Algorithms}
Malicious attacker is able to steal private information by using gradient attack algorithms \cite{DBLP:conf/nips/ZhuLH19, DBLP:conf/aistats/BagdasaryanVHES20, DBLP:conf/ccs/HitajAP17}. Therefore, clients can't transmit gradients but model parameters directly. FL server optimizes neural network by aggregating all parameters of clients (which is updated a few epochs locally) into global one. Typically, server aggregates model parameters collected from $K$ clients at round $i$ to update global weights $w^{i+1}$ following FedAvg \cite{DBLP:conf/aistats/McMahanMRHA17}:
    $$
    w^{i+1} = \sum_{k=1}^K \frac{n_k}{n}w_k^i
    $$
Under this setting, FL optimization still faces many challenges.
In data-center distributed machine learning, each computation node get its dataset from parameter server, which makes data distribution independently identically distribution (I.I.D) across nodes. However, data in FL clients can be Non-I.I.D in many ways \cite{hsieh2020non}, which leads to inferior robustness and slow convergence.  

Plenty of federated optimization algorithms are proposed to overcome data None-I.I.D problem. \cite{DBLP:conf/iclr/WangYSPK20, li2018federated, DBLP:conf/iclr/LiSBS20,  DBLP:conf/icml/KarimireddyKMRS20} try to learn a better shared federated model based on different aggregation strategies. FL Pensonalization \cite{DBLP:conf/iclr/ZhangSFYA21, DBLP:journals/corr/abs-2007-03797} aims to learn personalized model for every client. The combination of FL and other deep learning techniques, such as meta learning \cite{jiang2019improving}, transfer learning \cite{seo2020federated}, etc., are popular as well. To summarize, most optimization researches only relate to local training process on client and parameter aggregation process on FL server, which indicates that a flexible FL framework shall provide customizable interfaces for both local training design as well as server aggregation strategies.

\subsection{Communication and Compression}
Bandwidth problem is bottleneck of large-scale distributed training, and it becomes even worse when distributed training is performed in FL. Thus, deploying communication compression strategy is necessary, especially in cross-device setting. Two common-used and low resource-consumption compression methods as follows: \par \textbf{Quantization} \cite{alistarh2017qsgd, dettmers20158, bernstein2018signsgd} replaces each tensor with a lower precision one (e.g., float16 instead of float32), accomplishing the trade-off between precision and compression ratio.
\textbf{Sparsification} \cite{DBLP:conf/emnlp/AjiH17, lin2017deep, stich2018sparsified} selects a subset of tensors by appointed principle (e.g., Top-$k$ selection) to transmit. It can achieve at least 100$\times$ compression ratio. These two compression methods are model independent, which shows a flexible FL framework shall also provide model-independent compression module.

\subsection{Related work}
Several open-sources FL frameworks have been released. FATE\footnote{\url{https://fate.fedai.org/}} is a large federated secure computing framework. PaddleFL\footnote{\url{https://github.com/PaddlePaddle/PaddleFL}} and FedLearner\footnote{\url{https://github.com/bytedance/fedlearner}} are proposed by Baidu and Bytedance that support applications and deployment of FL system in application scenario. Frameworks above are industrial-oriented, focusing on real-life applications but not suitable for laboratory FL simulation. Rosetta \cite{Rosetta} and PySyft \cite{ryffel2018generic} mainly focus on secure multiparty computation of FL rather than algorithm and communication researches. TFF\footnote{\url{https://github.com/tensorflow/federated}} supports the simulation of FL training but executes only on a single machine. FedML \cite{chaoyanghe2020fedml} is a comprehensive FL framework that includes most research fields in FL. And Flower \cite{beutel2020flower} provides a FL communication framework supporting different deep learning framework (e.g., PyTorch, TensorFlow and MXNet). But they still hold varies of dependent libraries, which makes them heavy.

Different from frameworks above, \texttt{FedLab} is designed to be lightweight. It focuses on optimization effectiveness and communication efficiency for FL system simulation. We encourage users to build FL system following standard program pipeline and providing custom interfaces at the same time. Features of \texttt{FedLab} are further illustrated in the next section.

\begin{figure}
    \centering
    \includegraphics[width=0.8\textwidth]{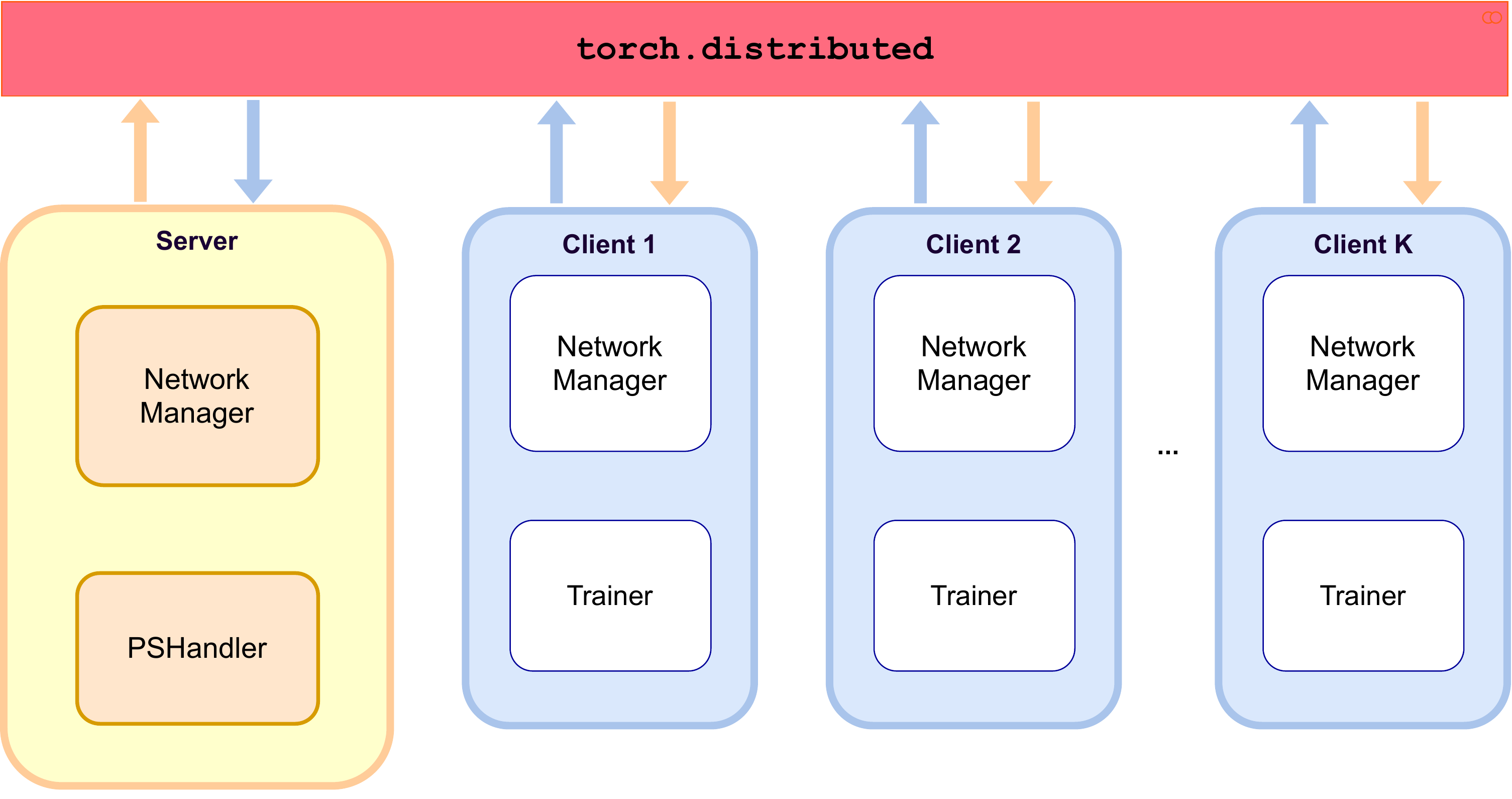}
    \caption{An overview of FedLab architecture. Two main roles in FedLab are define with two functional module: \texttt{NetworkManager} and \texttt{ParameterServerHandler/Trainer}. Communication backend is \texttt{torch.distributed} module.}
    \label{fig:overviw}
\end{figure}

\section{Framework Overview} \label{sec:overview}

In this section, we mainly illustrate architectural designs and detailed features in both communication efficiency and optimization effectiveness aspects. 
 \texttt{FedLab} provides two main roles in FL setting: Server and Client. Each Server/Client consists of two components called \texttt{NetworkManager} and \texttt{ParameterServerHandler}/\texttt{Trainer}. The overview of \texttt{FedLab}'s structure is shown in Figure \ref{fig:overviw}.

\texttt{NetworkManager} module manages message process task, which provides interfaces to customize communication agreements and compression algorithms. In section \ref{sec:com}, the details of communication module is demonstrated. \texttt{ParameterServerHandler}/\texttt{Trainer} takes charge of specific optimization algorithm design, and is illustrated in section \ref{sec:optim}. Finally, three deployment scenarios supported by \texttt{FedLab} are presented in section \ref{sec:deploy}. 

\subsection{Communication Efficiency}\label{sec:com}
In order to meet various requirements of FL network communication, \texttt{FedLab} implements \texttt{NetworkManager} to manage network topology, using \texttt{torch.distributed} as communication backend. \texttt{NetworkManager} is designed to be flexible in tensor agnostic, customization and scalability. Details of these features are stated below.

\begin{figure}
\centering  %图片全局居中
\subfigure[Synchronous]{
\label{fig.sychronous}
\includegraphics[width=0.6\textwidth]{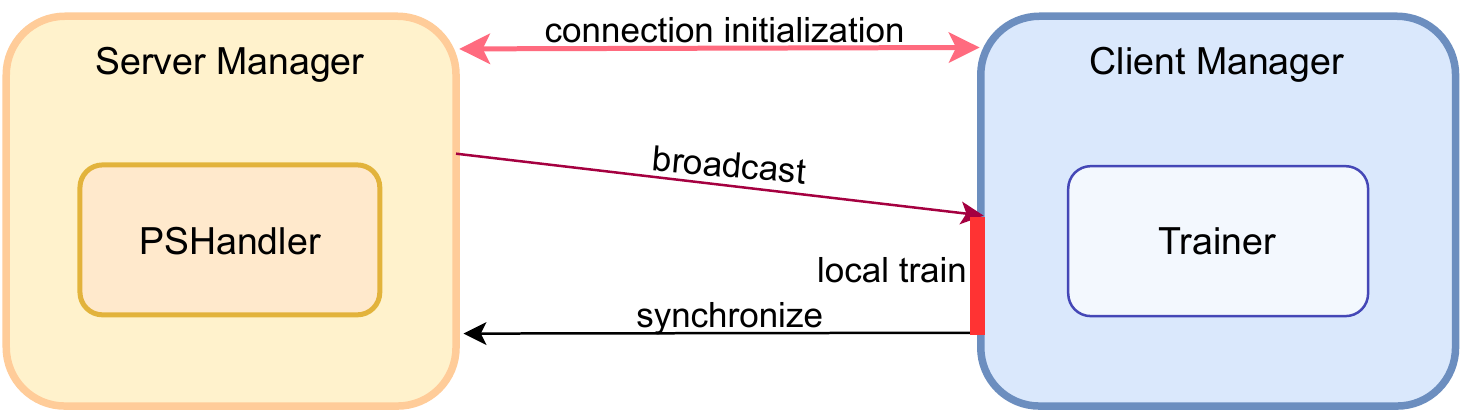}}

\subfigure[Asynchronous]{
\label{Fig.asychronous}
\includegraphics[width=0.6\textwidth]{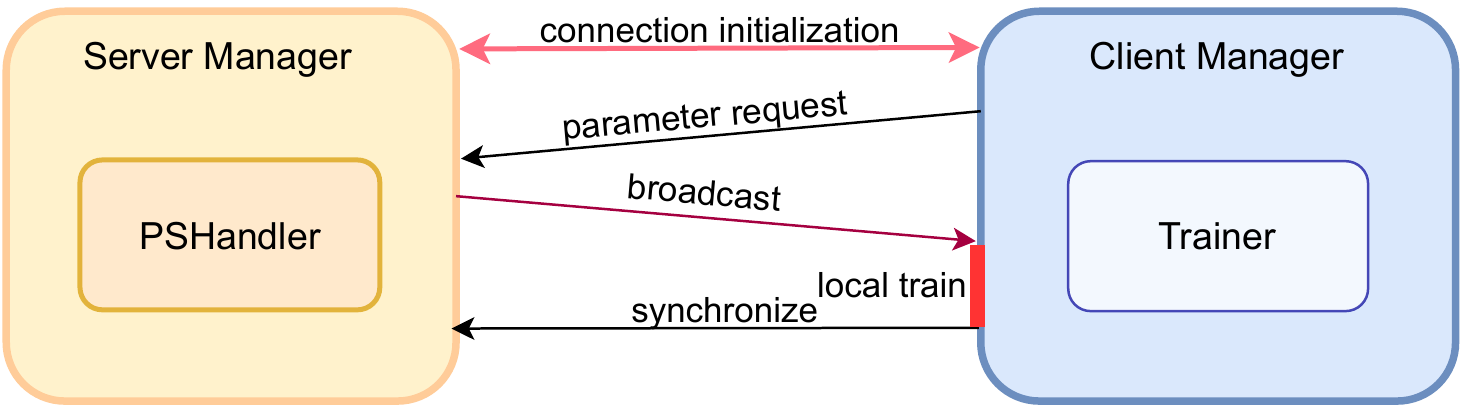}}

\caption{NetworkManager in FedLab}
\label{Fig.NetworkManager}
\end{figure}

\textbf{Tensor-based communication}. Inspired by the structure of network message, the basic communication element in \texttt{FedLab} is called \texttt{Package}, which contains \emph{header tensor} with necessary control information and \emph{content tensor} with packed tensor list. What's more, \texttt{PackageProcessor} in \texttt{NetworkManager} provides useful functions for packing up tensor list and restoring content to tensor list. In this way, the details of \texttt{Package} are blocked from users. Besides, \texttt{Package} is represented by a one-dimension tenser (vector), which is compatible with interfaces of PyTorch precisely. 

\textbf{Communication Agreement Customizable}. Communication agreements can be explained by following questions: What contents to send? How does client or server react after receiving message? Flexibility of communication module is given by \texttt{NetworkManager} module, which offers users the interfaces of customizing communication protocol. User can define additional information exchange, and control information flow for advanced algorithm development. 

\textbf{Communication Pattern}. Synchronous and Asynchronous communication patterns are implemented according to Federated Optimization algorithms. Specifically for figure \ref{fig.sychronous}, One round of synchronous communication flow can be describe as follows:
\begin{enumerate}[label=\arabic*)]
    \item \emph{Initialization}. Server and Clients initialize network connection.
    \item \emph{Sampling}. Server selects subset of clients to join current round of FL by broadcasting global model to them.
    \item \emph{Synchronization}. Client starts it local train process after receiving global model. Then, every Client sends needed information including local model to Server.
    \item \emph{Aggregation}. Finally, Server collects all information from Clients and performs aggregation.
\end{enumerate}

Differently, in asynchronous communication, every client communicate with server asynchronously. A FL training round is begin with a parameter request from client. Besides, server update global model every time it receives a synchronization upload. Details are shown in figure \ref{Fig.asychronous}.

\textbf{Scheduler}. \emph{Cross-silo} and \emph{Cross-device} \cite{kairouz2019advances} are the common FL settings. Cross-silo FL system usually has 2 - 100 clients which with large bandwidth and powerful computing resources. In contrary, cross-device scenario indicates that more clients (up to $10^{10}$) but less resources (power, bandwidth) with each client. Since there are needs of simulating more than 100 of clients, we designed message forward module \texttt{Scheduler} to extend the scalability of \texttt{FedLab}. Firstly, \texttt{Scheduler} is able to connect machines in different LAN(Local Area Network). What's more, users can overwrite the work flow of \texttt{Scheduler} to achieve hierarchical communication pattern. The usage of \texttt{Scheduler} will be further illustrated in section \ref{sec:deploy}.\par

\subsection{Optimization Effectiveness}\label{sec:optim}

%Gradient attack algorithms\cite{DBLP:conf/nips/ZhuLH19, DBLP:conf/aistats/BagdasaryanVHES20, DBLP:conf/ccs/HitajAP17} draw privacy information from gradient. FL model optimization algorithm by aggregate all parameters of clients(which is updated a few epochs locally) into global one. 
Optimization module in \texttt{FedLab} achieves "high-cohesion and low-coupling", which means this module can be used independently just like LEGOs bricks. To be more specific, \texttt{ParameterServerHandler} and \texttt{Trainer} is executable without \texttt{NetworkManager}. Besides, \texttt{FedLab} does not provide high level APIs, but prepares the necessary implementation tools for developers, reflecting the flexibility of framework.
In this section, some key features of \texttt{FedLab} for standard FL optimization are illustrated.

\begin{figure}[htbp]
\centering  %图片全局居中

\subfigure[Standalone]{
\label{fig.SerialTrainer}
\includegraphics[width=0.45\textwidth]{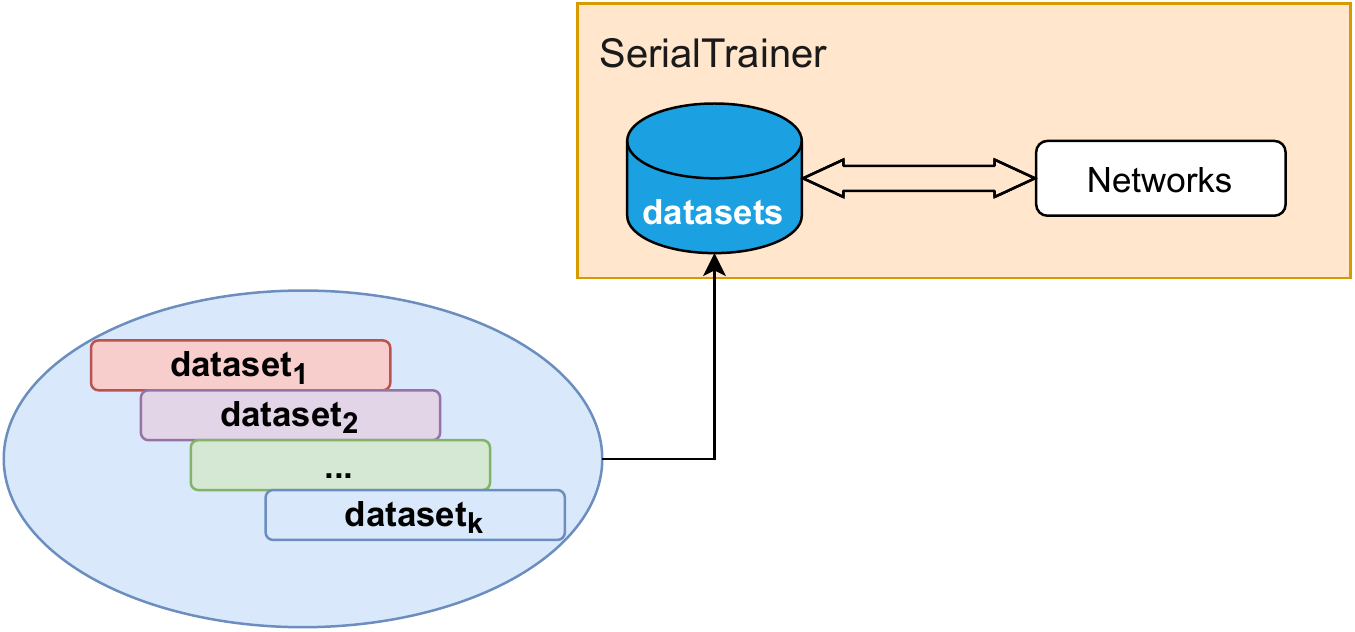}}\hspace{5mm}
\subfigure[Cross-process]{
\label{fig.multi_process}
\includegraphics[width=0.45\textwidth]{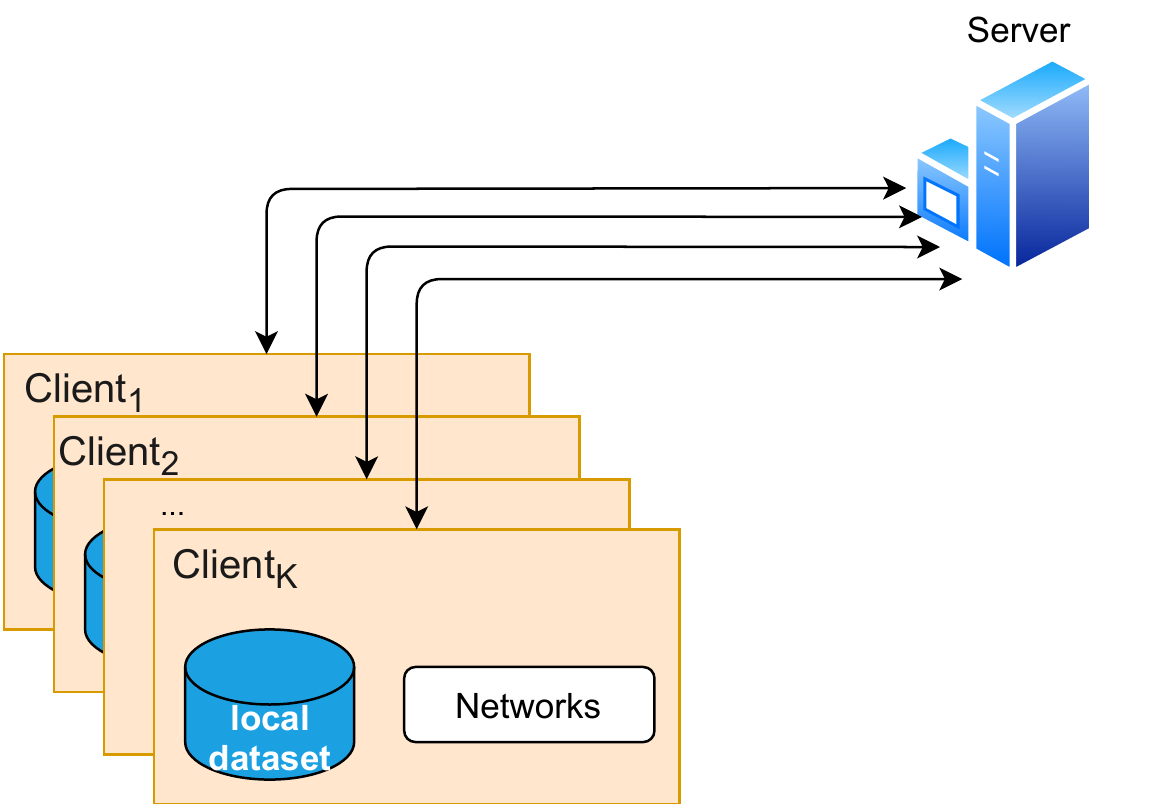}}

\subfigure[Hierarchical]{
\label{hierarchical}
\includegraphics[width=0.7\textwidth]{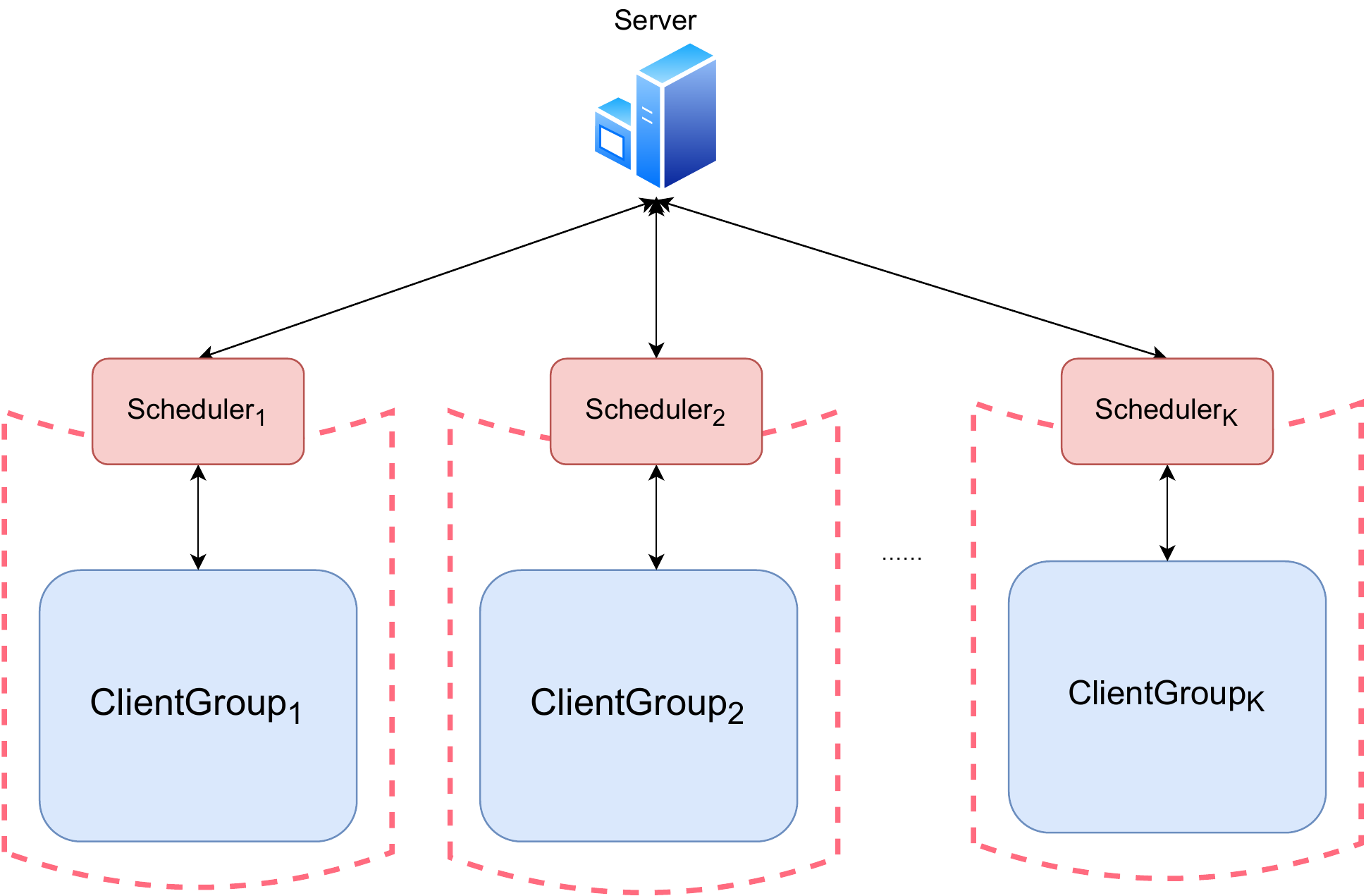}}

\caption{Supported Deployment Scenario in FedLab}
\label{Fig.deployment}
\end{figure}

\textbf{Aggregation}. 
\texttt{Trainer}/\texttt{ParameterServerHandler} in \texttt{FedLab} is corresponding with Client/Server optimization process. We encourage standard optimization implementation paradigm for both Client and Server: \texttt{Trainer} manages local dataset and performs PyTorch training process. \texttt{ParameterServerHandler} is implementation of parameter aggregation. 
In \texttt{FedLab}, \texttt{ClientSGDTrainer} is a standard implementation of \texttt{Trainer} for users. Additionally, we provides standard demos of \texttt{ParameterServerHandler} with different aggregation algorithms, such as FedAvg \cite{DBLP:conf/aistats/McMahanMRHA17} and FedAsgd \cite{DBLP:journals/corr/abs-1903-03934}.\par

\textbf{Data Partition}. 
In practice, Non-I.I.D datasets are not always accessible for researchers due to privacy restrictions. Thus, researchers tend to manually create Non-I.I.D data partition in experiment environment. For instance, FedAvg \cite{DBLP:conf/aistats/McMahanMRHA17} sorts the MNIST dataset by digit label, and divides it into 2000 shards of size 300 to create pathological Non-I.I.D partition over clients. Also, current FL researches handling non-IID problems tend to design very specific non-IID scenarios rather than standard and systematic partition schemes \cite{DBLP:journals/corr/abs-2102-02079}. Therefore, \texttt{FedLab} offers users a series of data partition functions, as well as built-in data partition schemes for some datasets based on design of NIID-bench \cite{DBLP:journals/corr/abs-2102-02079} and \cite{DBLP:conf/iclr/AcarZNMWS21}. What's more, \texttt{FedLab} provides PyTorch version of LEAF \cite{DBLP:journals/corr/abs-1812-01097} (a Non-I.I.D partitioned FL datasets baseline).\par

\subsection{Deployment Scenarios}\label{sec:deploy}
\texttt{FedLab} encapsulates the network interface of \texttt{torch.distributed} module, providing stable end-to-end \texttt{tensor} transmission for FL simulation. Furthermore, we implement a scalable version of \texttt{NetworkManager}, called \texttt{Scheduler}, to ensure the flexibility of network topology and the scalability of the system.
Different deployment scenarios of \texttt{FedLab} correspond to different experimental conditions, for scalability and flexibility.

\textbf{Standalone}. 
\texttt{FedLab} implements \texttt{SerialTrainer} for FL simulation in single process. \texttt{SerialTrainer} allows user to simulate a FL system with multiple clients, only with limited computation resources. However, it consumes more time to finish the whole FL experiment since the clients' real execution is one by one in serial. It is designed for simulation with limited computation resources. The paradigm of \texttt{SerialTrainer} is shown in figure \ref{fig.SerialTrainer}.
 
\textbf{Cross-process}. 
\texttt{FedLab} also supports cross-process FL simulation that's shown in figure \ref{fig.multi_process}. In practice, each role of \texttt{FedLab} is represented by single system process. FL system simulation can be executed on multiple machines with correct network topology configuration. More flexibly in parallel, \texttt{SerialTrainer} is able to replace the regular \texttt{Trainer}. In this way, machine with more computation resources can be assigned with more workload of simulating. The limitation of this scenario is that all machines must be in the same network (LAN or WAN).

\textbf{Hierarchical}. 
Users can break the limitation of \textbf{cross-process} by using \texttt{Scheduler} to build client groups (a subset of clients sharing the same \texttt{Scheduler}), as depicted in figure \ref{hierarchical}. Server can communicate with client in LAN indirectly. A hierarchical FL system with $K$ client groups as depicted in figure \ref{hierarchical} can be easily formed using \texttt{FedLab}. More importantly, \texttt{Scheduler} is customizable for users. It can be applied for aggregating parameters from client group as a middle-server to share the communication and computation load of server. This design is for the scalability of framework in both computation and communication.

\section{Pipeline and Examples}

The pipeline of building a FL system with \texttt{FedLab} includes two parts. The first part is definition of communication agreements. The prototype of synchronous and asynchronous communication patterns have been implemented for users. With effortless modification on \texttt{NetworkManager} of client and server, users can fulfill the agreements as their will.
The second part is \texttt{ParameterServerHandler} module of server and \texttt{Trainer} module of client, which represents FL optimization process. High level parameter aggregation algorithm and communication is available for server as well. In short, customizable interfaces and tools in \texttt{FedLab} support users to implement these two parts very quickly. We show the example implementation of FedAvg to demonstrate \texttt{FedLab} API's simplicity. 

Core code of client is shown below:
\begin{lstlisting}[language=Python, morekeywords={torch, ClientSGDTrainer, DistNetwork, ClientPassiveManager, nn , CrossEntropyLoss}]
model = ResNet()
optimizer = torch.optim.SGD(model.parameters(), lr=args.lr, momentum=0.9)
criterion = nn.CrossEntropyLoss()
trainloader, testloader = get_dataset(args)

handler = ClientSGDTrainer(model, trainloader, epoch=args.epoch, optimizer=optimizer, criterion=criterion, cuda=args.cuda)
network = DistNetwork(address=(args.server_ip, args.server_port),
                      world_size=args.world_size,
                      rank=args.local_rank)
                      
manager = ClientPassiveManager(handler=handler, network=network)
manager.run()
\end{lstlisting}
Code from line 1 to line 5 is the standard pipeline of training a neural network with PyTorch. From line 6 to the end is the usage of \texttt{FedLab}. In this example, \texttt{FedLab} provides high level API of network communication which allow users define network topology easily (line 7-11) and standard network training process (line 6).\par
FL server is also easily implemented in a couple lines of code:
\begin{lstlisting}[language=Python, morekeywords={torch, SyncParameterServerHandler, ServerSynchronousManager, DistNetwork}]
model = ResNet()
ps = SyncParameterServerHandler(model, client_num_in_total=args.world_size-1)

network = DistNetwork(address=(args.server_ip, args.server_port), 
                      world_size=args.world_size, 
                      rank=0)
manager = ServerSynchronousManager(handler=ps, network=network)
    
manager.run()
\end{lstlisting}
Code in line 2 defines the \texttt{ParameterServerHandler} with FedAvg algorithm. Codes from line 4 to 7 define the \texttt{NetworkManager} of server.

\section{Development}
For continuous maintainence of \texttt{FedLab}, we establish a open-source group on GitHub. The framework will be further developed publicly through GitHub, in which we can track issues of bug reports, feature requests and usage questions. We use continuous integration (CI) to ensure robust of package. What's more, comprehensive and elaborate documentation is developed using popular Sphinx Python documentation generator and published on \codeword{fedlab.readthedocs.io}.

\section{Summary and Future Work}
In this paper, a flexible and lightweight FL framework \texttt{FedLab} is proposed. \texttt{FedLab} provides common-used FL communication patterns and optimization algorithms modules with both high-level API and open interfaces for standardized FL simulation. For easy usage and continuous maintainence, we build a open-source group to accept contributions and issues on GitHub with necessary configurations.

In the future, we will keep developing \texttt{FedLab}. Specifically, our plan includes but not limited to the following aspects:
\begin{itemize}
    \item \textbf{Releasing research results}. We will use \texttt{FedLab} to explore our current and future ideas about optimization and communication. We will release those implementations on this framework in the future.
    \item \textbf{Providing more implementations}. Many excellent works are developed by different computation platform. Inconsistent implementations are not beneficial for the development of community. We plan to re-implement them with \texttt{FedLab} to provide more standard FL implementations.
    \item \textbf{Adding functional modules}. In the aspect of communication module, complicate network topology is under development. Besides, convenient network configuration script will be presented soon. Modules, which supporting other machine learning technique such as Unsupervised Learning, Semi-supervised Learning, Transfer Learning, etc,, are in schedule.
    
\end{itemize}

\bibliographystyle{unsrt}  
\bibliography{references}


\end{document}